\documentclass{article} 
\usepackage{iclr2019_conference,times}

\usepackage{booktabs} 
\usepackage[utf8]{inputenc} 
\usepackage[T1]{fontenc}    
\usepackage{hyperref}       
\usepackage{url}            
\usepackage{booktabs}       
\usepackage{amsfonts}       
\usepackage{nicefrac}       
\usepackage{microtype}      
\usepackage{color}
\usepackage{amsthm}
\usepackage{subcaption}
\usepackage{enumitem}
\usepackage{footnote}
\usepackage{algpseudocode}
\usepackage{algorithm}
\usepackage{amsmath}

\makesavenoteenv{table}

\usepackage{amsmath, amsfonts, amssymb, xspace, color, amsbsy}
\usepackage{tikz, graphicx}
\usetikzlibrary{bayesnet}
\usetikzlibrary{positioning}

\newcommand{\hide}[1]{} 

\newcommand{\eg}{\emph{e.g.}\xspace} 

\newcommand{\Sum}{\sum\limits} 

\newcommand{\norm}[1]{\mathopen\| #1 \mathclose\|}			

\def \x {\mathbf{x}}
\def \y {\mathbf{y}}

\newcommand{\tweet}{\texttt{Tweet}\xspace}
\newcommand{\blog}{\texttt{BlogCatalog}\xspace}
\newcommand{\yt}{\texttt{YouTube}\xspace}

\title{The Importance of Norm Regularization in Linear Graph Embedding: Theoretical Analysis and Empirical Demonstration}

\iclrfinalcopy
\author{Yihan Gao, Chao Zhang, Jian Peng \& Aditya Parameswaran \\
Department of Computer Science\\
University of Illinois at Urbana-Champaign\\
Urbana, IL 61801, USA \\
\texttt{\{ygao34,czhang82,jianpeng,adityagp\}@illinois.edu} \\
}

\newtheorem{claim}{Claim}
\newtheorem{theorem}{Theorem}

\newcommand{\eat}[1]{}

\newcommand{\techreport}[1]{}

\begin{document}

\maketitle

\begin{abstract}

  Learning distributed representations for nodes in graphs is a crucial primitive in 
  network analysis with a wide spectrum of applications. Linear graph embedding methods learn such
  representations by optimizing the likelihood of both positive and negative edges
  while constraining the dimension of the embedding vectors. We argue that the generalization 
  performance of these methods  is not due to the dimensionality constraint as commonly believed, 
  but rather the small norm of embedding vectors. Both theoretical and empirical evidence are provided 
  to support this argument: (a) we prove that the generalization error of these methods 
  can be bounded by limiting the norm of vectors, regardless of the embedding dimension; (b) 
  we show that the generalization performance of linear graph embedding methods is 
  correlated with the norm of embedding vectors, which is small due to the early stopping 
  of SGD and the vanishing gradients. We performed extensive experiments to validate 
  our analysis and showcased the importance of proper norm regularization in practice.

\end{abstract}

\section{Introduction}
\label{sect:intro}

Graphs have long been considered as one of the most fundamental structures that
can naturally represent interactions between numerous real-life objects (\eg, the
Web, social networks, protein-protein interaction networks). Graph embedding, whose 
goal is to learn distributed representations for nodes while preserving the structure
of the given graph, is a fundamental problem in network analysis that underpins many
applications. A handful of graph embedding techniques have been
proposed in recent years \citep{Perozzi2014, Tang2015, GroverL16}, along with impressive
results in applications like link prediction, text classification
\citep{Tang2015a}, and gene function prediction \citep{wang2015exploiting}.

Linear graph embedding methods preserve graph structures by converting 
the inner products of the node embeddings into probability distributions with a softmax 
function \citep{Perozzi2014, Tang2015, GroverL16}. Since the exact softmax objective is 
computationally expensive to optimize, the negative sampling technique \citep{mikolov2013distributed}
is often used in these methods: instead of optimizing the softmax objective function,
we try to maximize the probability of positive instances while minimizing the probability of 
some randomly sampled negative instances. It has been shown that by 
using this negative sampling technique, these graph embedding methods are essentially 
computing a factorization of the adjacency (or proximity) matrix of graph
\citep{levy2014neural}. Hence, it is commonly believed that the key to the 
generalization performance of these methods is the dimensionality constraint.

However, in this paper we argue that \emph{the key factor to the good generalization of
these embedding methods is not the dimensionality constraint, but rather the small 
norm of embedding vectors}. We provide both theoretical and empirical evidence to 
support this argument:
\begin{itemize}[leftmargin=15pt]
	\vspace{-5pt}
	\item Theoretically, we analyze the generalization error of two linear graph embedding 
	hypothesis spaces (restricting embedding dimension/norm), and show that only the 
	norm-restricted hypothesis class can theoretically guarantee good generalization 
	in typical parameter settings.
	\item Empirically, we show that the success of existing linear graph embedding 
	methods~\citep{Perozzi2014, Tang2015, GroverL16} are due to the early stopping of stochastic 
	gradient descent (SGD), which implicitly restricts the norm of embedding vectors.
	Furthermore, with prolonged SGD execution and no proper norm regularization, 
	the embedding vectors can severely overfit the training data.
\end{itemize}

\subsection*{Paper Outline}

The rest of this paper is organized as follows. In Section~\ref{sect:prelim}, we review the 
definition of graph embedding problem and the general framework of linear graph embedding. In Section~\ref{sect:norm}, we present both theoretical and empirical evidence to support our argument 
that the generalization of embedding vectors is determined by their norm. In Section~\ref{sect:method}, 
we present additional experimental results for a hinge-loss linear graph embedding variant, which
further support our argument. In Section~\ref{sect:discussion}, we discuss the new insights 
that we gained from previous results. Finally in Section~\ref{sect:conclusion}, we conclude our
paper. Details of the experiment settings, algorithm pseudo-codes, theorem proofs and the discussion of other related work can all be found in the appendix.

\section{Preliminiaries}
\label{sect:prelim}

\subsection{The Graph Embedding Problem}

We consider a graph $G = (V, E)$, where $V$ is the set of nodes in $G$, and $E$
is the set of edges between the nodes in $V$.  For any two nodes $u, v \in V$,
an edge $(u, v) \in E$ if $u$ and $v$ are connected, and we assume all
edges are unweighted and undirected for simplicity\footnote{All linear graph embedding 
methods discussed in this paper can be generalized to weighted case by multiplying
the weight to the corresponding loss function of each edge. The directed case is usually 
handled by associating each node with two embedding vectors for incoming and outgoing edges respectively, which is equivalent as learning embedding on a transformed undirected 
bipartite graph.}. The task of graph embedding is to learn a $D$-dimensional
vector representation $\x_u$ for each node $u \in V$ such that the structure of $G$ 
can be maximally preserved. These embedding vectors can then be used as features for 
subsequent applications (e.g., node label classification or link prediction).

\subsection{The Linear Graph Embedding Framework}

Linear graph embedding~\citep{Tang2015, GroverL16} is one of the two major approaches
for computing graph embeddings~\footnote{The other major approach is to use deep neural 
network structure to compute the embedding vectors, see the discussion of other related 
works in the appendix for details.}. These methods use the inner products 
of embedding vectors to capture the likelihood of edge existence, and are appealing to 
practitioners due to their simplicity and good empirical performance. 
Formally, given a node $u$ and its neighborhood $N_+(u)$ \footnote{Note that 
$N_+(u)$ can be either the set of direct neighbors in the original graph $G$~\citep{Tang2015}, 
or an expanded neighborhood based on measures like random walk~\citep{GroverL16}.}, the 
probability of observing node $v$ being a neighbor of $u$ is defined as:
$$ p(v | u) = \frac{\exp(\x_u^T \x_v)}{\Sum_{k \in V} \exp(\x_u^T \x_k)}. $$

By minimizing the KL-divergence between the embedding-based distribution and the 
actual neighborhood distribution, the overall objective function is equivalent to:
$$ L = - \Sum_{u \in E} \sum_{v \in N_+(u)} \log p(v | u) $$

Unfortunately, it is quite problematic to optimize this objective function directly, as the 
softmax term involves normalizing over all vertices. To address this issue, the negative 
sampling~\citep{mikolov2013distributed} technique is used to avoid computing gradients 
over the full softmax function. Intuitively, the negative sampling technique can be viewed as 
randomly selecting a set of nodes $N_-(u)$ that are not connected to each node $u$ as its \emph{negative neighbors}. The embedding vectors are then learned by minimizing the 
following objective function instead:
\begin{equation}\label{eqn_ns}
L = - \Sum_u \Sum_{v \in N_+(u)} \log \sigma(\x_u^T \x_v) - \Sum_u \Sum_{v \in N_-(u)} \kappa 
\frac{|N_+(u)|}{|N_-(u)|} \log \sigma(-\x_u^T \x_v).
\end{equation}

\subsection{The Matrix Factorization Interpretation}\label{sec_mf}

Although the embedding vectors learned through negative sampling do have good empirical
performance, there is very few theoretical analysis of such technique that explains
the good empirical performance. The most well-known analysis of negative sampling was 
done by~\citet{levy2014neural}, which claims that the embedding vectors are approximating 
a low-rank factorization of the PMI (Pointwise Mutual Information) matrix.

More specifically, the key discovery of \citet{levy2014neural} is that when the embedding
dimension is large enough, the optimal solution to Eqn~(\ref{eqn_ns}) recovers exactly 
the PMI matrix (up to a shifted constant, assuming the asymptotic case where $N_-(u) = V$ 
for all $u \in V$):
$$ \forall u, v, \x_u^T \x_v = \log (\frac{|E| \cdot 1_{(u,v) \in E}}{|N_+(u)| |N_+(v)|}) - \log \kappa$$

Based on this result, \citet{levy2014neural}~suggest that optimizing Eqn~(\ref{eqn_ns}) under
the dimensionality constraint is equivalent as computing a low-rank factorization of the 
shifted PMI matrix. This is currently the mainstream opinion regarding the intuition behind 
negative sampling. Although Levy and Goldberg only analyzed negative sampling
in the context of word embedding, it is commonly believed that the same conclusion also 
holds for graph embedding~\citep{qiu2018network}.
\vspace{-10pt}
\section{The Importance of Norm Regularization}
\label{sect:norm}

As explained in Section~\ref{sec_mf}, it is commonly believed that linear grpah embedding methods 
are approximating a low-rank factorization of PMI matrices.  As such, people often 
deem the dimensionality constraint of embedding vectors as the key factor to good
generalization~\citep{Tang2015, GroverL16}. However, due to the sparsity of real-world 
networks, the explanation of Levy \& Goldberg is actually very counter-intuitive in the graph embedding 
setting: the average node degree usually only ranges from 10 to 100, which is much less than
the typical value of embedding dimension (usually in the range of $100 \sim 400$). Essentially, 
this means that in the context of graph embedding, the total number of free parameters is 
larger than the total number of training data points, which makes it intuitively 
very unlikely that the negative sampling model (i.e., Eqn~(\ref{eqn_ns})) can 
inherently guarantee the generalization of embedding vectors in such scenario, and it 
is much more plausible if the observed good empirical performance is due to some other reason.

In this paper, we provide a different explanation to the good empirical performance of 
linear graph embedding methods: we argue that the good generalization of linear graph embedding
vectors is due to their small norm, which is in turn caused by the vanishing gradients during 
the stochastic gradient descent (SGD) optimization procedure. We provide the following evidence 
to support this argument:
\begin{itemize}
	\item In Section~\ref{sect:theory}, we theoretically analyze the generalization error 
	of two linear graph embedding variants: one has the standard dimensionality 
	constraints, while the other restricts the vector norms. 
	Our analysis shows that:
	\begin{itemize}
		\item The embedding vectors can generalize well to unseen data if their average
		squared $l_2$ norm is small, and this is always true regardless of the embedding 
		dimension choice.
		\item Without norm regularization, the embedding vectors can severely overfit the
		training data if the embedding dimension is larger than the average node degree.
	\end{itemize}
	\item In Section~\ref{sect:sgd}, we provide empirical evidence that the generalization of 
	linear graph embedding is determined by vector norm instead of embedding dimension. 
	We show that:
	\begin{itemize}
	\item In practice, the average norm of the embedding vectors is small due 
	to the early stopping of SGD and the vanishing gradients.
	\item The generalization performance of embedding vectors starts to drop when
	the average norm of embedding vectors gets large.
	\item The dimensionality constraint is only helpful when the embedding dimension is very small (around $5 \sim 10$) and there is no norm regularization.
	\end{itemize}	
\end{itemize}

\subsection{Generalization Analysis of Two Linear Graph Embedding Variants}\label{sect:theory}

In this section, we present a generalization error analysis of linear graph embedding 
based on the uniform convergence framework~\citep{bartlett2002rademacher}, which bounds the 
maximum difference between the training and generalization error over the entire hypothesis space. 
We assume the following statistical model for graph generation: 
there exists an unknown probability distribution $\mathcal{Q}$ over the
Cartesian product $V \times U$ of two vertex sets $V$ and $U$. Each sample $(a,b)$ 
from $\mathcal{Q}$ denotes an edge connecting $a \in V$ and $b \in U$.The set of (positive)
training edges $E_+$ consists of the first $m$ i.i.d. samples from the distribution 
$\mathcal{Q}$, and the negative edge set $E_-$ consists of i.i.d. samples from the 
uniform distribution $\mathcal{U}$ over $V \times U$. The goal is to use these 
samples to learn a model that generalizes well to the underlying distribution 
$\mathcal{Q}$. We allow either $V = U$ for homogeneous graphs or $V \cap U = 
\emptyset$ for bipartite graphs. 

Denote $E_\pm = \{(a,b,+1) : (a,b) \in E_+\} \cup \{(a,b,-1) : (a,b) \in E_-\}$ to be 
the collection of all training data, and we assume that data points in $E_\pm$ are 
actually sampled from a combined distribution $\mathcal{P}$ over $V \times U \times
\{\pm 1\}$ that generates both positive and negative edges. Using the above notations, 
the training error $\mathcal{L}_t(\x)$ and generalization error $\mathcal{L}_g(\x)$ 
of embedding $\x: (U \cup V) \rightarrow \mathbb{R}^D$ are defined 
as follows:
\begin{align*}
\mathcal{L}_t(\x) = \frac{1}{|E_\pm|} \sum_{(a,b,y) \in E_\pm} - \log \sigma(y \x_a^T \x_b) \qquad
\mathcal{L}_g(\x) = - \mathbb{E}_{(a,b,y) \sim \mathcal{P}} \log \sigma(y \x_a^T \x_b)
\end{align*}

In the uniform convergence framework, we try to prove the following statement:
$$ \mathbf{Pr}(\sup_{\x \in \mathcal{H}} (\mathcal{L}_g(\x) - \mathcal{L}_t(\x)) \leq \epsilon) \geq 1 -\delta $$
which bounds the maximum difference between $\mathcal{L}_t(\x)$ and $\mathcal{L}_g(\x)$ over all possible embeddings $\x$ in the hypothesis space $\mathcal{H}$. 
If the above uniform convergence statement is true, then minimizing the training error
$\mathcal{L}_t(\x)$ would naturally lead to small generalization error $\mathcal{L}_g(\x)$ 
with high probability. 

Now we present our first technical result, which follows the above framework and bounds the generalization error of linear graph embedding methods with norm constraints:

\begin{theorem}\label{thm:bipartite} [Generalization of Linear Graph Embedding with Norm Constraints]
	Let $E_\pm = \{(a_1, b_1, y_1), (a_2, b_2, y_2), \ldots, (a_{m+m'}, b_{m+m'}, y_{m+m'})\}$ 
	be i.i.d.~samples from a distribution $\mathcal{P}$ over $V \times U \times \{\pm 1\}$. 
	Let $\x : (U \cup V) \rightarrow \mathbb{R}^D$ to be the embedding for nodes in the graph. 
	Then for any bounded $1$-Lipschitz loss function $l : \mathbb{R} \rightarrow [0, B]$ 
	and $C_U, C_V > 0$, with probability $1 - \delta$ (over the sampling of $E_\pm$), the following 
	inequality holds
	\begin{align*}
	& \mathbb{E}_{(a,b,y) \sim \mathcal{P}} l(y \x_a^T \x_b) \leq \frac{1}{m+m'} \sum_{i=1}^{m+m'} 
	l(y_i \x_{a_i}^T \x_{b_i}) + \frac{2}{m+m'} \mathbb{E}_\sigma ||A_\sigma||_2 \sqrt{C_U C_V} + 4 B \sqrt{\frac{2 \ln (4 / \delta)}{m+m'}}
	\end{align*}
	for all embeddings $\x$ satisfying
	$$ \sum_{u \in U} \norm{\x_u}^2 \leq C_U, \sum_{v \in V} \norm{\x_v}^2 \leq C_V $$
	where $||A_\sigma||_2$ is the spectral norm of the randomized adjacency
	matrix $A_\sigma$ defined as follows:
	\begin{align*}
	A_\sigma(i,j) = \left\{
	\begin{array}{lr}
	\sigma_{ij} & \exists y, (u_i, v_j, y) \in E_\pm \\
	0 & \forall y, (u_i,v_j,y) \notin E_\pm \\
	\end{array}
	\right.
	\end{align*}
	in which $\sigma_{ij}$ are i.i.d. Rademacher random variables.
\end{theorem}

The proof can be found in the appendix. Intuitively, this theorem states that the size of the 
training dataset (i.e., the total number of edges $|E_\pm| = m + m'$) required for learning a 
norm constrained graph embedding is about $O(C ||A_{\sigma}||_2)$, where $C$ is the sum of 
squared norm of all embedding vectors and $\mathbb{E}_\sigma ||A_{\sigma}||_2$ is the expected
spectral norm of the randomized adjacency matrix $A_{\sigma}$ (defined in the theorem). 
Assuming that the average squared norm of embedding vectors is $O(1)$, then $C$ scales as 
$O(|V| + |U|)$, and a rough estimate of $\mathbb{E}_\sigma||A_{\sigma}||_2$ puts it 
around $O(\sqrt{ \frac{|E_\pm| \ln (|U|+|V|)}{|U|+|V|}})$\footnote{This estimation is computed on 
a Erdos–Renyi style random graph, details can be found in the appendix.}. Based on these 
estimates, we see that the generalization error gap of norm constrained embeddings scales as
$O(d^{-0.5} (\ln n)^{0.5})$\footnote{We suspect that a more accurate estimation could remove 
the $(\ln n)^{0.5}$ factor.}, where $d$ is the average node degree and $n$ is the total
number of vertices.

On the other hand, if we restrict only the embedding dimension (i.e., no norm regularization 
on embedding vectors), and the embedding dimension is larger than the average degree of the 
graph, then it is possible for the embedding vectors to severely overfit the training data.
The following example demonstrates this possibility on a $d$-regular graph, in which 
the embedding vectors can always achieve zero training error even when the edge labels 
are randomly placed:
\begin{claim}\label{clm:low_dim}
	Let $G = (V,E)$ be a $d$-regular graph with $n$ vertices and $m = nd / 2$ labeled edges 
	(with labels $y_i \in \{\pm 1\}$):
	$$V = \{v_1, \ldots, v_n\} \quad E = \{(a_1, b_1, y_1), \ldots, (a_m, b_m, y_m)\}$$
	Then for each of the $2^m$ possible combination of labels, there always exists a $d$-dimensional 
	embedding $\x: V \rightarrow \mathbb{R}^d$ that achieves perfect classification accuracy on
	the randomized training dataset:
	\begin{align*}
		& \forall (y_1, \ldots, y_m) \in \{\pm 1\}^m, \exists \x: V \rightarrow \mathbb{R}^d \\
		\textbf{s.t.} \quad & \forall i \in \{1, \ldots, m\}, \quad y_i \x_{a_i}^T \x_{b_i} > 1
	\end{align*}
\end{claim}

The proof can be found in the appendix. In other words, without norm regularization, the number 
of training samples required for learning $D$-dimensional embedding vectors is at least 
$\Omega(nD)$. Considering the fact that many large-scale graphs are sparse (with average 
degree $< 20$) and the default embedding dimension commonly ranges from $100$ to $400$, 
it is highly unlikely that the the dimensionality constraint 
by itself could lead to good generalization performance. 


\subsection{Empirical Evidence on the Cause of Generalization}\label{sect:sgd}

In this section, we present several sets of experimental results for the standard linear graph 
embedding, which collectively suggest that the generalization of these methods are actually 
determined by vector norm instead of embedding dimension. 

\noindent \textbf{Experiment Setting:} We use stochastic gradient descent (SGD) to minimize the 
following objective:
$$ L = - \lambda_{+1} \sum_{(u,v) \in E_+} \log \sigma(\x_u^T \x_v) \nonumber - \lambda_{-1} \sum_{(u,v) \in E_-} \log \sigma(- \x_u^T \x_v) + \lambda_r \sum_{v \in V} ||\x_v||_2^2 $$
Here $E_+$ is the set of edges in the training graph, and $E_-$ is the set of negative edges with 
both ends sampled uniformly from all vertices. The SGD learning rate is standard: $\gamma_t = (t +
c)^{-1/2}$. Three different datasets are used in the experiments: \tweet, \blog 
and \yt, and their details can be found in the appendix.

\begin{figure}[h]
	\centering
	\begin{subfigure}[b]{0.3\textwidth}
		\centering
		\includegraphics[width = 4cm]{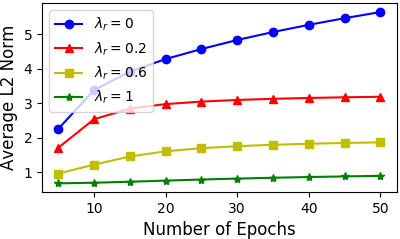}
		\caption{\tweet}\label{fig:epoch_norm_twt}
	\end{subfigure}
	~ 
	\begin{subfigure}[b]{0.3\textwidth}
		\centering
		\includegraphics[width = 4cm]{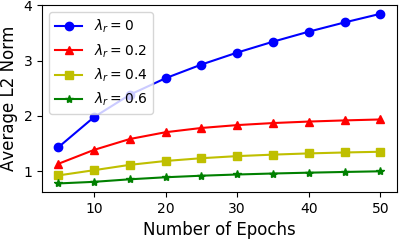}
		\caption{\blog}\label{fig:epoch_norm_blg}
	\end{subfigure}
	~ 
	\begin{subfigure}[b]{0.3\textwidth}
		\centering
		\includegraphics[width = 4cm]{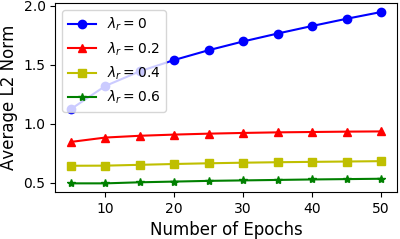}
		\caption{\yt}\label{fig:epoch_norm_yt}
	\end{subfigure}
	\vspace{-5pt}
	\caption{Average $l_2$ Norm during SGD Optimization}
	\label{fig:norm_epoch}
	\vspace{-5pt}		
\end{figure}

\noindent \textbf{SGD Optimization Results in Small Vector Norm:} Figure~\ref{fig:norm_epoch} 
shows the average $l_2$ norm of the embedding vectors during the first 50 SGD epochs 
(with varying value of $\lambda_r$). As we can see, the average norm of embedding 
vectors increases consistently after each epoch, but the increase rate gets slower as time 
progresses. In practice, the SGD procedure is often stopped after $10\sim50$ epochs 
(especially for large scale graphs with millions of vertices\footnote{Each epoch of SGD 
has time complexity $O(|E|D)$, where $D$ is the embedding dimensionality (usually around 
$100$). Therefore in large scale graphs, even a single epoch would require billions of 
floating point operations.}), and the relatively early stopping time would 
naturally result in small vector norm.

\begin{figure}[h]
	\centering
	\begin{subfigure}[b]{0.3\textwidth}
		\centering
		\includegraphics[width = 4cm]{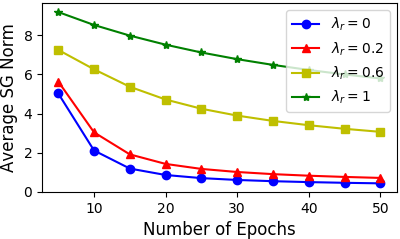}
		\caption{\tweet}\label{fig:epoch_grad_twt}
	\end{subfigure}
	~ 
	\begin{subfigure}[b]{0.3\textwidth}
		\centering
		\includegraphics[width = 4cm]{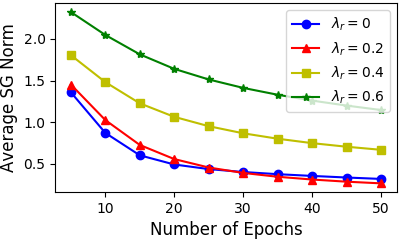}
		\caption{\blog}\label{fig:epoch_grad_blg}
	\end{subfigure}
	~ 
	\begin{subfigure}[b]{0.3\textwidth}
		\centering
		\includegraphics[width = 4cm]{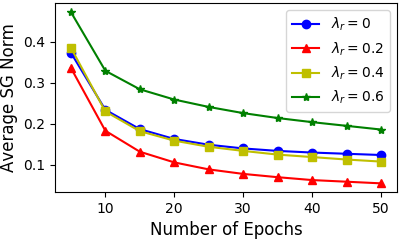}
		\caption{\yt}\label{fig:epoch_grad_yt}
	\end{subfigure}
	\vspace{-5pt}
	\caption{Average Norm of Stochastic Gradients during SGD Optimization}
	\label{fig:grad_epoch}
	\vspace{-5pt}	
\end{figure}

\noindent \textbf{The Vanishing Gradients:} Figure~\ref{fig:grad_epoch} shows the average $l_2$ norm 
of the stochastic gradients $\partial L/\partial \x_u$ during the first 50 SGD epochs:
\begin{equation}\label{eqn_sg}
\frac{\partial L}{\partial \x_u} = - \sum_{v \in N_+(u)} \sigma(-\x_u^T \x_v) \x_v + \sum_{v \in N_-(u)} \sigma(\x_u^T \x_v) \x_v + 2 \lambda_r \x_u
\end{equation}
From the figure, we can see that the stochastic gradients become smaller during the later 
stage of SGD, which is consistent with our earlier observation in Figure~\ref{fig:norm_epoch}. 
This phenomenon can be intuitively explained as follows: after a few SGD epochs, most of the 
training data points have already been well fitted by the embedding vectors, which means that 
most of the coefficients $\sigma(\pm \x_u^T \x_v)$ in Eqn~(\ref{eqn_sg}) will be close to $0$ 
afterwards, and as a result the stochastic gradients will be small in the following epochs. 

\begin{figure}[h]
	\centering
	\begin{subfigure}[b]{0.45\textwidth}
		\centering
		\includegraphics[width = 5cm]{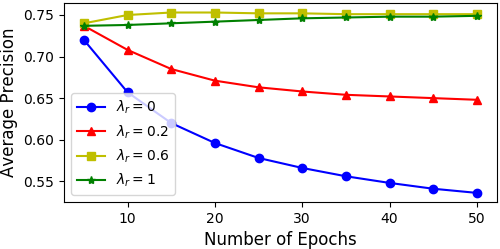}
		\caption{\tweet(AP)}\label{fig:epoch_ap_twt}
	\end{subfigure}
	~ 
	\begin{subfigure}[b]{0.45\textwidth}
		\centering
		\includegraphics[width = 5cm]{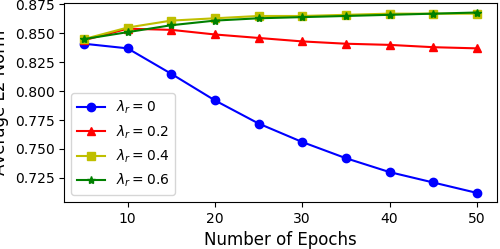}
		\caption{\blog(AP)}\label{fig:epoch_ap_blg}
	\end{subfigure}

	\begin{subfigure}[b]{0.45\textwidth}
		\centering
		\includegraphics[width = 5cm]{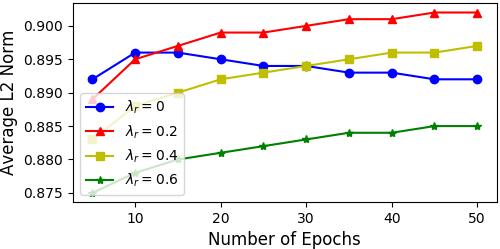}
		\caption{\yt(AP)}\label{fig:epoch_ap_yt}
	\end{subfigure}
	~ 
	\begin{subfigure}[b]{0.45\textwidth}
		\centering
		\includegraphics[width = 5cm]{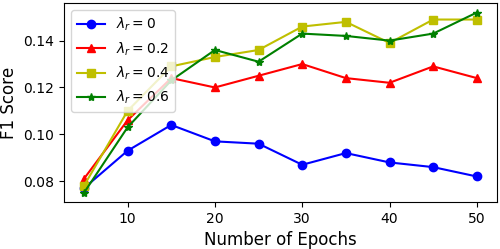}
		\caption{\blog(F1)}\label{fig:epoch_f1_blg}
	\end{subfigure}
	\vspace{-5pt}
	\caption{Generalization Performance During SGD}
	\label{fig:gen_epoch}
	\vspace{-10pt}	
\end{figure}

\noindent \textbf{Regularization and Early Stopping:} Figure~\ref{fig:gen_epoch} 
shows the generalization performance of embedding vectors during the first 50 SGD epochs, 
in which we depicts the resulting average precision (AP) score\footnote{Average Precision (AP) 
evaluates the performance on ranking problems: we first compute the precision and recall 
value at every position in the ranked sequence, and then view the precision $p(r)$ as a function 
of recall $r$. The average precision is then computed as $\textit{AveP} = \int_0^1 p(r) dr$.} 
for link prediction and F1 score for node label classification. As we can see, the generalization 
performance of embedding vectors starts to drop after $5\sim20$ epochs when $\lambda_r$ is small, 
indicating that they are overfitting the training dataset afterwards. The generalization 
performance is worst near the end of SGD execution when $\lambda_r = 0$, which coincides 
with the fact that embedding vectors in that case also have the largest norm among all settings. 
Thus, Figure~\ref{fig:gen_epoch} and Figure~\ref{fig:norm_epoch} collectively suggest that the 
generalization of linear graph embedding is determined by vector norm.

\begin{figure}[h]
	\vspace{-10pt}
	\centering
	\begin{subfigure}[b]{0.45\textwidth}
		\centering
		\includegraphics[width = 5cm]{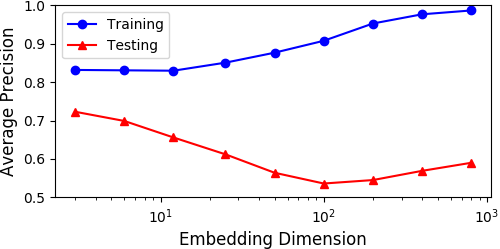}
		\caption{$\lambda_r = 0$}\label{fig:l0_dim_twt}
	\end{subfigure}
	~ 
	\begin{subfigure}[b]{0.45\textwidth}
		\centering
		\includegraphics[width = 5cm]{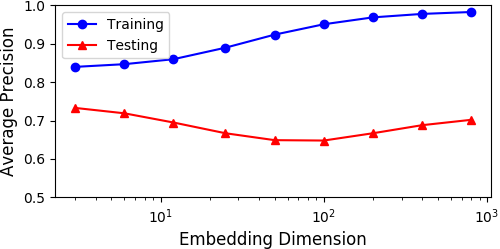}
		\caption{$\lambda_r = 0.2$}\label{fig:l2_dim_blg}
	\end{subfigure}
	
	\begin{subfigure}[b]{0.45\textwidth}
		\centering
		\includegraphics[width = 5cm]{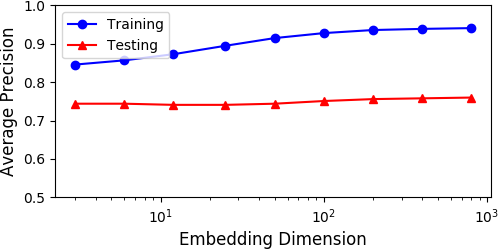}
		\caption{$\lambda_r = 0.6$}\label{fig:l6_dim_yt}
	\end{subfigure}
	~ 
	\begin{subfigure}[b]{0.45\textwidth}
		\centering
		\includegraphics[width = 5cm]{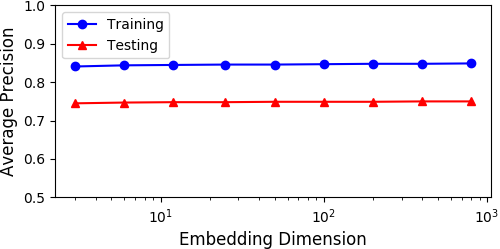}
		\caption{$\lambda_r = 1$}\label{fig:l1_dim_blg}
	\end{subfigure}
	\vspace{-5pt}
	\caption{Impact of Embedding Dimension}
	\vspace{-10pt}
	\label{fig:AP_dim}
\end{figure}

\noindent \textbf{Impact of Embedding Dimension Choice:} Figure~\ref{fig:AP_dim} shows the 
generalization AP score on \tweet dataset with varying value of $\lambda_r$ and embedding 
dimension $D$. As we can see in Figure~\ref{fig:AP_dim}, without any norm regularization 
($\lambda_r = 0$), the embedding vectors will overfit the training dataset for any $D$ 
greater than $10$, which is consistent with our analysis in Claim~\ref{clm:low_dim}. On 
the other hand, with larger $\lambda_r$, the impact of embedding dimension choice is 
significantly less noticeable, indicating that the primary factor for generalization 
is the vector norm in such scenarios.

\section{Demonstrating the Importance of Norm Regularization via Hinge-Loss Linear Graph Embedding}
\label{sect:method}

In this section, we present the experimental results for a non-standard linear graph embedding 
formulation, which optimizes the following objective:
\begin{align}\label{eqn:hinge_obj}
L = & \lambda_{+1} \sum_{(u,v) \in E_+} h(\x_u^T \x_v)  + \lambda_{-1} \sum_{(u,v) \in E_-} h(-\x_u^T \x_v) + \frac{\lambda_r}{2} \sum_{v \in V} ||\x_v||_2^2
\end{align}
By replacing logistic loss with hinge-loss, it is now possible to apply the dual
coordinate descent (DCD) method~\citep{HsiehCLKS08} for optimization, which circumvents 
the issue of vanishing gradients in SGD, allowing us to directly observe the impact of 
norm regularization. More specifically, consider all terms in Eqn~(\ref{eqn:hinge_obj}) 
that are relevant to a particular vertex $u$:
\begin{equation}\label{eqn:obj}
L(u) = \Sum_{(\x_i, y_i) \in D} \frac{\lambda_{y_i}}{\lambda_r} \max(1 - y_i \x_u^T \x_i, 0) + \frac{1}{2} \norm{\x_u}^2.
\end{equation}
in which we defined $D = \{(\x_v, +1): v \in N_+(u)\} \cup \{(\x_k, -1): k \in N_-(u)\}$. 
Since Eqn~(\ref{eqn:obj}) takes the same form as a soft-margin linear SVM objective,
with $\x_u$ being the linear coefficients and $(\x_i, y_i)$ being training data,
it allows us to use any SVM solver to optimize Eqn~(\ref{eqn:obj}), and then apply it 
asynchronously on the graph vertices to update their embeddings. The pseudo-code for the 
optimization procedure using DCD can be found in the appendix.

\begin{figure}[h]
	\centering
	\begin{subfigure}[b]{0.3\textwidth}
		\centering
		\includegraphics[width = 4cm]{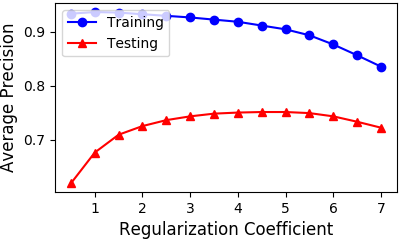}
		\caption{\tweet}\label{fig:lr_ap_twt}
	\end{subfigure}
	~ 
	\begin{subfigure}[b]{0.3\textwidth}
		\centering
		\includegraphics[width = 4cm]{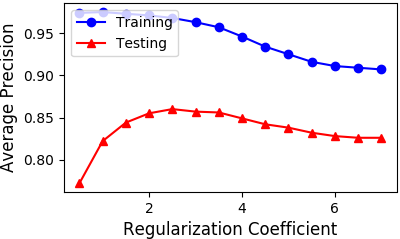}
		\caption{\blog}\label{fig:lr_ap_blg}
	\end{subfigure}
	~ 
	\begin{subfigure}[b]{0.3\textwidth}
		\centering
		\includegraphics[width = 4cm]{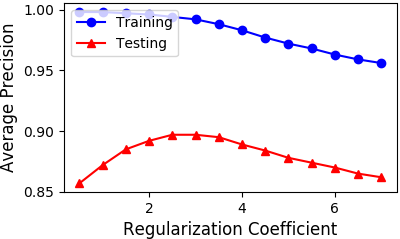}
		\caption{\yt}\label{fig:lr_ap_yt}
	\end{subfigure}
	\vspace{-5pt}
	\caption{Generalization Average Precision with Varying $\lambda_r$ ($D = 100$)}
	\label{fig:AP_reg}
	\vspace{-5pt}	
\end{figure}

\noindent \textbf{Impact of Regularization Coefficient:} Figure~\ref{fig:AP_reg} shows the
generalization performance of embedding vectors obtained from DCD procedure ($\sim 20$ epochs).
As we can see, the quality of embeddings vectors is very bad when $\lambda_r \approx 0$, 
indicating that proper norm regularization is necessary for generalization. The value of 
$\lambda_r$ also affects the gap between training and testing performance, which is
consistent with our analysis that $\lambda_r$ controls the model capacity of 
linear graph embedding.

\begin{figure}[h]
	\vspace{-10pt}
	\centering
	\begin{subfigure}[b]{0.3\textwidth}
		\centering
		\includegraphics[width = 4cm]{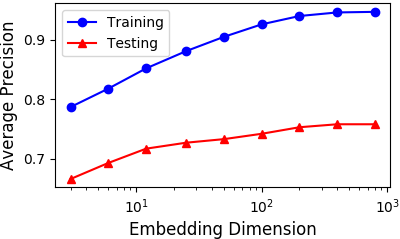}
		\caption{\tweet}\label{fig:dim_ap_twt}
	\end{subfigure}
	~ 
	\begin{subfigure}[b]{0.3\textwidth}
		\centering
		\includegraphics[width = 4cm]{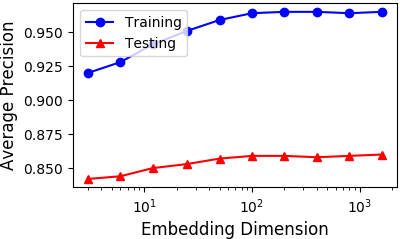}
		\caption{\blog}\label{fig:dim_ap_blg}
	\end{subfigure}
	~ 
	\begin{subfigure}[b]{0.3\textwidth}
		\centering
		\includegraphics[width = 4cm]{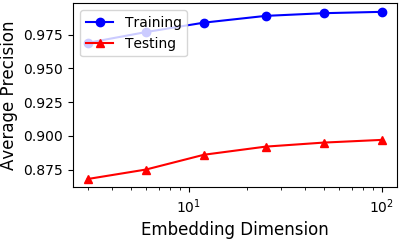}
		\caption{\yt}\label{fig:dim_ap_yt}
	\end{subfigure}
	\vspace{-5pt}
	\caption{Generalization Average Precision with Varying $D$ ($\lambda_r = 3$)}
	\label{fig:dim_2}
	\vspace{-10pt}	
\end{figure}

\noindent \textbf{Impact of Embedding Dimension Choice:} The choice of embedding dimension $D$
 on the other hand is not very impactful as demonstrated in Figure~\ref{fig:dim_2}: as long 
 as $D$ is reasonably large ($\geq 30$), the exact choice has very little effect on 
the generalization performance. Even with extremely large embedding dimension setting $(D 
= 1600)$. These results are consistent with our theory that the generalization of linear graph 
embedding is primarily determined by the norm constraints.

\section{Discussion}
\label{sect:discussion}

So far, we have seen many pieces of evidence supporting our argument, suggesting that the 
generalization of embedding vectors in linear graph embedding is determined by the vector norm. 
Intuitively, it means that these embedding methods are trying to embed the vertices onto a 
small sphere centered around the origin point. The radius of the sphere controls the model 
capacity, and choosing proper embedding dimension allows us to control the trade-off 
between the expressive power of the model and the computation efficiency. 

Note that the connection between norm regularization and generalization performance is actually 
very intuitive. To see this, let us consider the semantic meaning of embedding vectors: the
probability of any particular edge $(u, v)$ being positive is equal to
$$ \mathbf{Pr}(y=1|u,v) = \sigma(\x_u^T \x_v) = \sigma(\frac{\x_u^T \x_v}{||\x_u||_2 ||\x_v||_2} ||\x_u||_2 ||\x_v||_2) $$
As we can see, this probability value is determined by three factors: 
\begin{itemize}
	\item $\x_u^T \x_v / (||\x_u||_2 ||\x_v||_2)$, the cosine similarity between $\x_u$ and $\x_v$,
	evaluates the degree of agreement between the directions of $\x_u$ and $\x_v$.
	\item $||\x_u||_2$ and $||\x_v||_2$ on the other hand, reflects the degree of confidence we have
	regarding the embedding vectors of $u$ and $v$.
\end{itemize}
Therefore, by restricting the norm of embedding vectors, we are limiting the confidence level that 
we have regarding the embedding vectors, which is indeed intuitively helpful for preventing overfitting.  

It is worth noting that our results in this paper do not invalidate the analysis of 
\citet{levy2014neural}, but rather clarifies on some key points: as pointed out by 
\citet{levy2014neural}, linear graph embedding methods are indeed approximating the factorization
of PMI matrices. However, as we have seen in this paper, the embedding vectors are primarily 
constrained by their norm instead of embedding dimension, which implies that the resulting
factorization is not really a standard low-rank one, but rather a low-norm factorization:
$$ \x_u^T \x_v \approx \textit{PMI}(u,v) \quad \textit{s.t.} \quad \sum_u ||\x_u||_2^2 \leq C $$
The low-norm factorization represents an interesting alternative to the standard low-rank
factorization, for which our current understanding of such factorization is still very limited. 
Given the empirical success of linear graph embedding methods, it would be really helpful if we 
can have a more in-depth analysis of such factorization, to deepen our understanding and potentially inspire new algorithms.

\vspace{-5pt}
\section{Conclusion}\label{sect:conclusion}
\vspace{-5pt}

We have shown that the generalization of linear graph embedding methods are 
not determined by the dimensionality constraint but rather the norm of embedding vectors. 
We proved that limiting the norm of embedding vectors would lead to good 
generalization, and showed that the generalization of existing linear graph 
embedding methods is due to the early stopping of SGD and vanishing gradients. We 
experimentally investigated the impact embedding dimension choice, and demonstrated
that such choice only matters when there is no norm regularization. In most cases,
the best generalization performance is obtained by choosing the optimal value for the
norm regularization coefficient, and in such case the impact of embedding dimension case
is negligible. Our findings combined with the analysis of \citet{levy2014neural}
suggest that linear graph embedding methods are probably computing a low-norm
factorization of the PMI matrix, which is an interesting alternative to the standard
low-rank factorization and calls for further study.





\bibliography{cited}
\bibliographystyle{iclr2019_conference}

\newpage
\section*{Appendix}

\subsection*{Datasets and Experimental Protocols}

We use the following three datasets in our experiments:

\begin{itemize}
	\item \tweet is an undirected graph that encodes keyword co-occurrence relationships
	using Twitter data: we collected $\sim$1.1 million English tweets using Twitter's Streaming 
	API during 2014 August, and then extracted the most frequent 10,000 keywords as graph nodes 
	and their co-occurrences as edges. All nodes with more than 2,000 neighbors are removed as 
	stop words. There are 9,913 nodes and 681,188 edges in total.
	\item \blog~\citep{Zafarani+Liu:2009} is an undirected graph that contains the social 
	relationships between BlogCatalog users. It consists of 10,312 nodes and 333,983 undirected 
	edges, and each node belongs to one of the 39 groups.
	\item \yt~\citep{mislove-2007-socialnetworks} is a social network 
	among YouTube users.  It includes 500,000 nodes and 3,319,221 undirected edges\footnote{Available 
	at http://socialnetworks.mpi-sws.org/data-imc2007.html. We only used the subgraph induced by the 
	first 500,000 nodes since our machine doesn't have sufficient memory for training the whole graph. 
	The original graph is directed, but we treat it as undirected graph as in \citet{Tang2015}.}.
\end{itemize}

For each positive edge in training and testing datasets, we randomly sampled $4$ negative edges, 
which are used for learning the embedding vectors (in training dataset) and evaluating average 
precision (in testing dataset). In all experiments, $\lambda_+ = 1, \lambda_- = 0.03$, which 
achieves the optimal generalization performance according to cross-validation. All initial 
coordinates of embedding vectors are uniformly sampled form $[-0.1, 0.1]$.

\subsection*{Other Related Works}

In the early days of graph embedding research, graphs are only used as the intermediate data 
model for visualization \citep{kruskal1978multidimensional} or non-linear dimension reduction
\citep{tenenbaum2000global, belkin2001laplacian}. Typically, the first step is to construct
an affinity graph from the features of the data points, and then the low-dimensional embedding 
of graph vertices are computed by finding the eigenvectors of the affinity matrix. 

For more recent graph embedding techniques, apart from the linear graph embedding methods 
discussed in this paper, there are also methods \citep{wang2016structural, kipf2016semi,
hamilton2017inductive} that explore the option of using deep neural network structures 
to compute the embedding vectors. These methods typically try to learn a deep neural network 
model that takes the raw features of graph vertices to compute their low-dimensional embedding
vectors: SDNE \citep{wang2016structural} uses the adjacency list of vertices as input
to predict their Laplacian Eigenmaps; GCN \citep{kipf2016semi} aggregates the output 
of neighboring vertices in previous layer to serve as input to the current layer (hence
the name ``graph convolutional network''); GraphSage \citep{hamilton2017inductive} extends GCN
by allowing other forms of aggregator (i.e., in addition to the mean aggregator in GCN). 
Interestingly though, all these methods use only $2$ or $3$ neural network layers in 
their experiments, and there is also evidence suggesting that using higher number of layer
would result in worse generalization performance \citep{kipf2016semi}. Therefore, it still 
feels unclear to us whether the deep neural network structure is really helpful in the task of 
graph embedding.

Prior to our work, there are some existing research works suggesting that norm constrained 
graph embedding could generalize well. \citet{srebro2005maximum} studied the problem of 
computing norm constrained matrix factorization, and reported superior performance compared 
to the standard low-rank matrix factorization on several tasks. Given the connection between 
matrix factorization and linear graph embedding \citep{levy2014neural}, the results in our 
paper is not really that surprising.

\subsection*{Proof of Theorem~\ref{thm:bipartite}}

Since $E_\pm$ consists of i.i.d. samples from $\mathcal{P}$, by the uniform convergence theorem~\citep{bartlett2002rademacher, shalev2014understanding}, with probability $1 - \delta$:
\begin{align*}
& \forall \x, \quad \textit{s.t.} \quad \sum_{u \in U} \norm{\x_u}^2 \leq C_U, \sum_{v \in V} \norm{\x_v}^2 \leq C_V, \\
& \mathbb{E}_{(a,b,y) \sim \mathcal{P}} l(y \x_a^T \x_b) \leq \frac{1}{m+m'} \sum_{i=1}^{m+m'} 
l(y_i \x_{a_i}^T \x_{b_i}) + 2 \mathcal{R}(\mathcal{H}_{C_U, C_V}) + 4 B \sqrt{\frac{2 \ln (4 / \delta)}{m+m'}}
\end{align*}
where $\mathcal{H}_{C_U, C_V} = \{\x : \sum_{u \in U} \norm{\x_u}^2 \leq C_U, \sum_{v \in V} \norm{\x_v}^2 \leq C_V\}$ is the hypothesis set, and $\mathcal{R}(\mathcal{H}_{C_U, C_V})$ is the empirical Rademacher Complexity of $\mathcal{H}_{C_U, C_V}$, which has the following explicit form:
\begin{align*}
\mathcal{R}(\mathcal{H}_{C_U, C_V}) = \frac{1}{m + m'} \mathbb{E}_{\sigma_{a,b} \sim \{-1,1\}} \sup_{\x \in \mathcal{H}_{C_U, C_V}} \sum_i \sigma_{a_i, b_i} l(y_i \x_{a_i}^T \x_{b_i}) 
\end{align*} 
Here $\sigma_{a,b}$ are i.i.d. Rademacher random variables: $\mathbf{Pr}(\sigma_{a,b} = 1) 
= \mathbf{Pr}(\sigma_{a,b} = -1) = 0.5$. Since $l$ is $1$-Lipschitz, based on the Contraction
Lemma~\citep{shalev2014understanding}, we have:
\begin{align*}
\mathcal{R}(\mathcal{H}_{C_U, C_V}) \leq & \frac{1}{m + m'} \mathbb{E}_{\sigma_{a,b} \sim \{-1,1\}} \sup_{\x \in \mathcal{H}_{C_U, C_V}} \sum_i \sigma_{a_i, b_i} y_i \x_{a_i}^T \x_{b_i} \\
= & \frac{1}{m + m'} \mathbb{E}_{\sigma_{a,b} \sim \{-1,1\}} \sup_{\x \in \mathcal{H}_{C_U, C_V}} \sum_i \sigma_{a_i, b_i} \x_{a_i}^T \x_{b_i}
\end{align*}
Let us denote $X_U$ as the $|U|d$ dimensional vector obtained by concatenating all vectors $\x_u$,
and $X_V$ as the $|V|d$ dimensional vector obtained by concatenating all vectors $\x_v$:
$$ X_U = (\x_{u_1}, \x_{u_2}, \ldots, \x_{u_{|U|}}) \quad X_V = (\x_{v_1}, \x_{v_2}, \ldots, \x_{v_{|V|}}) $$
Then we have:
$$ ||X_U||_2 \leq \sqrt{C_U} \quad ||X_V||_2 \leq \sqrt{C_V} $$
The next step is to rewrite the term $\sum_i \sigma_{a_i, b_i} \x_{a_i}^T \x_{b_i}$ in matrix form:
\begin{align*}
  & \sup_{\x  \in \mathcal{H}_{C_U, C_V}} \sum_i \sigma_{a_i, b_i} \x_{a_i}^T \x_{b_i} \\
= & \sup_{||X_U||_2 \leq \sqrt{C_U}, ||X_V||_2 \leq \sqrt{C_V}} X_U^T [A_\sigma \otimes I_d] X_V \\
= & \sqrt{C_U} \norm{A_\sigma \otimes I_d}_2 \sqrt{C_V} 
\end{align*}
where $A \otimes B$ represents the Kronecker product of $A$ and $B$, and $||A||_2$ represents the spectral norm of $A$ (i.e., the largest singular value of $A$). 

Finally, since $||A \otimes I||_2 = ||A||_2$, we get the desired result in Theorem~\ref{thm:bipartite}.

\subsection*{Proof Sketch of Claim~\ref{clm:low_dim}}

We provide the sketch of a constructive proof here.

Firstly, we randomly initialize all embedding vectors. Then for each $v \in V$, consider all the relevant constraints to $\x_v$:
$$ \mathbf{C}_v = \{(a, b, y) \in E : a = v \text{ or } b = v\} $$
Since $G$ is $d$-regular, $|\mathbf{C}_v| \leq d$. Therefore, there always
exists vector $b \in \mathbb{R}^d$ satisfying the following $|C_v|$ constraints:
$$ \forall (a,b,y) \in C_v, y \x_a \x_b = 1 + \epsilon $$
as long as all the referenced embedding vectors are linearly independent. 

Choose any vector $b'$ in a small neighborhood of $b$ that is not the linear combination of any other 
$d - 1$ embedding vectors (this is always possible since the viable set is a $d$-dimensional sphere 
minus a finite number of $d-1$ dimensional subspaces), and set $\x_v \leftarrow b'$. 

Once we have repeated the above procedure for every node in $V$, it is easy to see that all 
the constraints $y\x_a^T \x_b : (a,b,y) \in E$ are now satisfied.

\subsection*{Rough Estimation of $||A_\sigma||_2$ on Erdos–Renyi Graph}

By the definition of spectral norm, $||A_\sigma||_2$ is equal to:
$$ ||A_\sigma||_2 = \sup_{||\x||_2 = ||\y||_2 = 1, \x, \y \in \mathbb{R}^n} \y^T A_\sigma \x $$
Note that,
$$ \y^T A_\sigma \x = \sum_{(i,j) \in E} \sigma_{ij} y_i x_j $$
Now let us assume that the graph $G$ is generated from a Erdos-Renyi model (i.e., the probability of any pair $u,v$ being directed connected is independent), then we have:
$$ \y^T A_\sigma \x = \sum_{i} \sum_{j} \sigma_{ij} e_{ij} y_i x_j $$
where $e_{ij}$ is the boolean random variable indicating whether $(i,j) \in E$.

By Central Limit Theorem,
$$ \sum_{i} \sum_{j} \sigma_{ij} e_{ij} y_i x_j \sim \mathcal{N}(0, \frac{m}{n^2})$$
where $m$ is the expected number of edges, and $n$ is the total number of vertices. Then we have, 
$$ \mathbf{Pr}(\y^T A_\sigma \x \geq t) \approx O(e^{-\frac{t^2 n^2}{2 m}}) $$
for all $||\x||_2 = ||\y||_2 = 1$. 

Now let $S$ be an $\epsilon$-net of the unit sphere in $n$ dimensional Euclidean space, which has roughly $O(\epsilon^{-n})$ total number of points. Consider any unit vector $\x, \y \in \mathbb{R}^n$, and let $\x_S, \y_S$ be the closest point of $\x, \y$ in $S$, then:
\begin{align*}
\y^T A_\sigma \x = & (\y_S + \y - \y_S)^T A_\sigma (\x_S + \x - \x_S) \\
= & \y_S^T A_\sigma \x_S + (\y - \y_S)^T A_\sigma \x_S + \y_S^T A_\sigma (\x - \x_S) + (\y - \y_S)^T A_\sigma (\x - \x_S) \\
\leq &  \y_S^T A_\sigma \x_S + 2 \epsilon n + \epsilon^2 n
\end{align*}
since $||A_\sigma|| \leq n$ is always true.

By union bound, the probability that at least one pair of $\x_S, \y_S \in S$ satisfying $\y_S^T A_\sigma \x_S \geq t$ is at most:
$$ \mathbf{Pr}(\exists \x_S, \y_S \in S : \y_S^T A_\sigma \x_S \geq t) \approx O(\epsilon^{-2n} e^{-\frac{t^2 n^2}{2 m}}) $$
Let $\epsilon = 1/n, t = \sqrt{8m \ln n/n}$, then the above inequality becomes:
$$ \mathbf{Pr}(\exists \x_S, \y_S \in S : \y_S^T A_\sigma \x_S \geq t) \approx O(e^{-n \ln n}) $$
Since $\forall \x_S, \y_S \in S, \y_S^T A_\sigma \x_S < t$ implies that
$$\sup_{||\x||_2 = ||\y||_2 = 1, \x, \y \in \mathbb{R}^n} \y^T A_\sigma \x < t + 2 \epsilon n + \epsilon^2 n$$
Therefore, we estimate $||A_{\sigma}||_2$ to be of order $O(\sqrt{m \ln n/n})$.

\subsection*{Pseudocode of Dual Coordinate Descent Algorithm}

Algorithm~\ref{alg_DCD} shows the full pseudo-code of the DCD method for optimizing the hinge-loss variant of linear graph embedding learning.

\begin{algorithm}[t]
	\footnotesize
	\caption{DCD Method for Hinge-Loss Linear Graph Embedding}
	\label{alg_DCD}
	\begin{algorithmic}[5]
		\Function{DcdUpdate}{$u, N_{+}(u), N_{-}(u)$}
		\State $D = \{(v, +1) : v \in N_+(u)\} \cup \{(v, -1) : v \in N_-(u)\} $
		\State $w \leftarrow \sum_{(v,s) \in D} \alpha_{uv} s \x_v$
		\For{$(v, s) \in D$}
			\State $G \leftarrow s w^T \x_v - 1$  
			\State $U \leftarrow \lambda_s / \lambda_r$
			\State $PG \leftarrow \left\{ \begin{array}{lr} 
			\min(G, 0) & \text{if } \alpha_{uv} = 0 \\
			\max(G, 0) & \text{if } \alpha_{uv} = U \\
			G &	\text{Otherwise}
			\end{array} \right.$
			\If{$PG \not= 0$}
				\State $Q = \x_v^T \x_v$
				\State $\bar{\alpha}_{uv} \leftarrow \alpha_{uv}$
				\State $\alpha_{uv} \leftarrow \min(\max(\alpha_{uv} - G / Q), 0, U)$
				\State $w \leftarrow w + (\alpha_{uv} - \bar{\alpha}_{uv}) s x_v$
			\EndIf
		\EndFor
		\State $\x_u \leftarrow w$
		\EndFunction
		\Function{Main}{$V,E_+,E_-,\lambda_+,\lambda_-,\lambda_r$}
		\State Randomly initialize $\x_v$ for all $v \in V$
		\State Initialize $\alpha_{uv} \leftarrow 0$ for all $(u, v) \in E_+ \cup E_-$
		\For{$t \in 1, \ldots, T$}
			\For{$u \in V$}
				\State $N_{+}(u) \leftarrow \{v \in V: (u, v) \in E_+\}$
				\State $N_{-}(u) \leftarrow \{v \in V: (u, v) \in E_-\}$
				\State DCDUpdate($u, N_{+}(u), N_{-}(u)$)
			\EndFor
		\EndFor		
		\EndFunction
	\end{algorithmic}
\end{algorithm}




\end{document}